\documentclass[usenames,dvipsnames,sigconf]{acmart}
\usepackage{amsmath}
\usepackage{amsfonts}
\usepackage{latexsym}
\usepackage{url}
\usepackage{mathtools}
\usepackage{mathdots}
\usepackage{color}
\usepackage{siunitx}
\usepackage{array}
\usepackage{tabularx}
\usepackage{makecell}
\usepackage[normalem]{ulem}  % Necessary for strikeouts!
\usepackage{booktabs}
\usepackage{multirow}
\usepackage{graphicx}
\usepackage{textcomp}
\usepackage{xcolor}
\usepackage{bbm}
\usepackage[ruled,vlined]{algorithm2e}
\usepackage{cleveref}
\useunder{\uline}{\ul}{}
\usepackage{pifont}% http://ctan.org/pkg/pifont
\newcommand{\cmark}{\ding{51}}%
\newcommand{\xmark}{\ding{55}}%
\newcommand{\newpara}[1]{\vspace{0.5em} \noindent \textbf{#1} \hspace{0.5em}}

\DeclareMathOperator*{\argmax}{arg\,max}
\DeclareMathOperator*{\argmin}{arg\,min}

\AtBeginDocument{%
  \providecommand\BibTeX{{%
    \normalfont B\kern-0.5em{\scshape i\kern-0.25em b}\kern-0.8em\TeX}}}

\setcopyright{iw3c2w3g}
\copyrightyear{2021}
\acmYear{2021}
\begin{document}

\title{Self-Supervised Bot Play for Conversational Recommendation with Justifications}

\author{Shuyang Li}
\affiliation{%
  \institution{UC San Diego}
}
\email{shl008@ucsd.edu}

\author{Bodhisattwa Prasad Majumder}
\affiliation{%
  \institution{UC San Diego}
}
\email{bmajumde@ucsd.edu}

\author{Julian McAuley}
\affiliation{%
  \institution{UC San Diego}
}
\email{jmcauley@ucsd.edu}

\renewcommand{\shortauthors}{Li, et al.}

\begin{abstract}
Conversational recommender systems offer the promise of interactive, engaging ways for users to find items they enjoy.
We seek to improve conversational recommendation via three dimensions:
1) We aim to mimic a common mode of human interaction for recommendation: experts justify their suggestions, a seeker explains why they don't like the item, and both parties iterate through the dialog to find a suitable item.
2) We leverage ideas from conversational critiquing to allow users to flexibly interact with natural language justifications by critiquing subjective aspects.
3) We adapt conversational recommendation to a wider range of domains where crowd-sourced ground truth dialogs are not available.
We develop a new two-part framework for training conversational recommender systems.
First, we train a recommender system to jointly suggest items and justify its reasoning with subjective aspects.
We then fine-tune this model to incorporate iterative user feedback via self-supervised bot-play.
Experiments on three real-world datasets demonstrate that our system can be applied to different recommendation models across diverse domains to achieve superior performance in conversational recommendation compared to state-of-the-art methods.
We also evaluate our model on human users, showing that systems trained under our framework provide more useful, helpful, and knowledgeable recommendations in warm- and cold-start settings..
\end{abstract}


\begin{teaserfigure}
    \centerline{\includegraphics[width=1.0\linewidth]{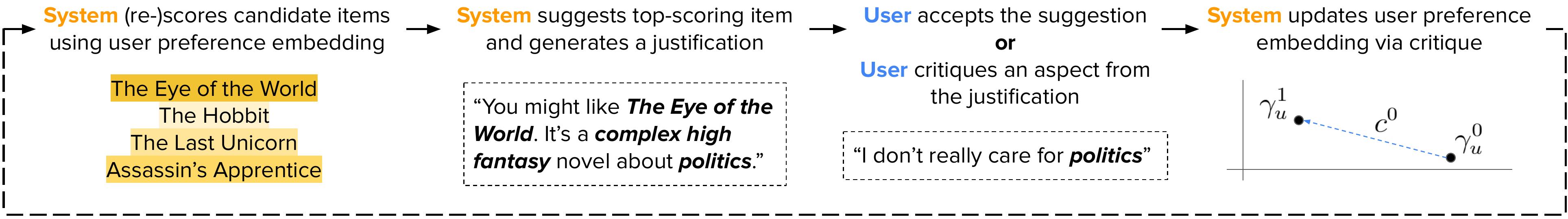}}
    \caption{Conversational critiquing workflow. The system scores candidate items and generates a justification for the top item. If the user rejects the suggestion and critiques an aspect, the system uses the critique to update the latent user representation.}
    \label{fig:pipeline}
\end{teaserfigure}

\maketitle

\section{Introduction}

Traditional recommender systems often return \textit{static} recommendations, with no way for users to meaningfully express their preferences and feedback.
However, interactivity and explainability can greatly affect a user's trust of and willingness to use a recommender system \cite{explanationgood,convgood}.
This is reflected in human conversations: experts justify their recommendations, customers critique suggestions, and both parties iterate through the conversation to arrive at a satisfactory item.

Early work on interactive recommender systems focused on iteratively presenting suggestions to the user based on simple ``like" and ``dislike" feedback on individual items \cite{img0,img1,seeker}.
Gradually, systems began to accommodate more fine-grained user feedback---critiquing fixed attributes of an item (e.g.~its color or brand) \cite{critsurvey}.
Recent models for conversational critiquing incorporate user feedback on subjective aspects (e.g.~taste and perception) \cite{WuCEVNCF,CEVAE,LLC}.
However, such methods are trained using a next-item recommendation objective, and perform poorly when engaging with users over multiple turns.

Another approach lies in training dialog agents to interact with the user over multiple turns \cite{moviedial1}.
While such models are able to generate convincing dialog in a vacuum, they require large corpora of transcripts from crowd-sourced recommendation games \cite{recdial,BotPlayRec}.
To create high-quality training dialogs, crowd-workers must be knowledgable about many items in the target domain---this expertise requirement limits data collection to a few common domains like movies.
Additionally, these dialog policies limit a user's freedom to interact with the system by asking yes/no questions about specific item attributes.

We thus desire a conversational recommender system that mimics characteristics of human interactions not yet captured by existing systems:
\begin{enumerate}
    \item It can justify suggestions made to the user;
    \item It updates suggestions based on user feedback about subjective aspects; and
    \item It can be trained using review data that is easily harvestable from arbitrary new domains.
\end{enumerate}

To accomplish these goals, we present a two-part framework to train conversational recommender systems.
Ours is the first framework, to our knowledge, that allows training of conversational recommender systems for multi-turn settings without the need for supervised dialog examples.
First, using a next-item recommendation task we learn to  encode historical user preferences and generate justifications for our suggestions via sets of subjective aspects.
We then fine-tune our trained model via multiple turns of bot-play in a recommendation game based on user reviews and simulated critiques.
We apply our framework to two base recommender systems (PLRec \cite{plrec} and BPR \cite{bpr}), and evaluate the resulting \textbf{PLRec-Bot} and \textbf{BPR-Bot} models on three large real-world recommendation datasets containing user reviews.
Our method reaches the target item with a higher success rate than state-of-the-art methods, and takes fewer turns to do so, on average.
We also conduct a study with real users, showing that it can effectively help users find desired items in real time, even in a cold-start setting.

% Our contributions
We summarize our main contributions as follows: 
1) We present a framework for training conversational recommender systems using bot-play on historical user reviews, without the need for large collections of human dialogs;
2) We apply our framework to two popular recommendation models (\textbf{BPR-Bot} and \textbf{PLRec-Bot}), with each showing superior or competitive performance in comparison to state-of-the-art recommendation and critiquing methods;
3) We demonstrate that our framework can be effectively combined with query refinement techniques to quickly suggest desired items.

% Prelims aka lit review?
\section{Related Work}

\begin{table}[t!]
\centering
\caption{Conversational critiquing systems (first section) are transcript-free but not equipped for multi-turn interactions. Dialog-based agents (second section) learn multi-turn interactions using large corpora of domain-specific dialog transcripts. Our framework allows us to train multi-turn interactive recommender systems without costly transcript data.}
\vspace{-0.8em}
\label{tab:lit-comparison}
\begin{tabular}{@{}lccc@{}}

\toprule
              & Justification & Multi-Turn & Transcript-Free \\ \midrule
LLC \cite{LLC}          &  \textcolor{red}{\xmark}             &  \textcolor{red}{\xmark}           & \textcolor{ForestGreen}{\cmark}               \\
CE-VAE \cite{CEVAE}        & \textcolor{ForestGreen}{\cmark}             &     \textcolor{red}{\xmark}       & \textcolor{ForestGreen}{\cmark}               \\
M\&M VAE \cite{MMVAE}      & \textcolor{ForestGreen}{\cmark}             &       \textcolor{red}{\xmark}     & \textcolor{ForestGreen}{\cmark}               \\ \midrule
\citet{recdial}       &      \textcolor{red}{\xmark}         & \textcolor{ForestGreen}{\cmark}          &      \textcolor{red}{\xmark}           \\
\citet{BotPlayRec}     & \textcolor{ForestGreen}{\cmark}             & \textcolor{ForestGreen}{\cmark}          &        \textcolor{red}{\xmark}         \\
\citet{KG-ConvRec}    & \textcolor{ForestGreen}{\cmark}             & \textcolor{ForestGreen}{\cmark}          &          \textcolor{red}{\xmark}       \\ \midrule
\textbf{Ours} & \textcolor{ForestGreen}{\cmark}             & \textcolor{ForestGreen}{\cmark}          & \textcolor{ForestGreen}{\cmark}               \\ \bottomrule
\end{tabular}
\end{table}

\subsection{Recommendation with Justification}
\label{litreview-just}

Users prefer recommendations that they perceive to be transparent or justified \cite{transparency,explanationgood}.
Some early recommender systems presented the objective attributes of suggested items to users \cite{tag1,tag2}, but did not attempt to personalize the justifications.
Another line of work considered the problem of generating natural language explanations of recommendations.
\citet{beeradvocate} extract key aspects from the text of user reviews using topic extraction.
Such justifications have been expanded into full sentences based on aspects of interest---constructed via template-filling \cite{templatejust} or recurrent language models \cite{jianmojust}.
Due to the unstructured nature of these justifications, however, sentence-level justifications have not been used for iteratively refining recommendations.
In this work, justifications take the form of specific aspects that a user is interested in (e.g.~that a song is \textit{poetic}) \cite{beeradvocate,WuCEVNCF,TREC}.
In \Cref{sec:aspect-extraction} we describe how we extract such aspects from user reviews in large recommendation datasets.

\begin{table}[t!]
\centering
\caption{Notation used in this paper.}
\vspace{-0.4em}
\label{tab:notation}
\begin{tabular}{@{}ll@{}}
\toprule
Notation & Description                             \\ \midrule
$\mathbf{K}^{U} \in \mathbb{R}^{|U| \times |K|}$ & \begin{tabular}[c]{@{}l@{}}User aspect frequency; $\mathbf{k}^{U}_{u,a}$ is how often\\ user $u$ mentioned aspect $a$ in their reviews.\end{tabular} \\
$\mathbf{K}^{I} \in \mathbbm{1}^{|I| \times |K|}$  & \begin{tabular}[c]{@{}l@{}}Binary matrix;  $\mathbf{k}^{I}_{i,a}$ is 1 if and only if aspect $a$ \\was used in any review of item $i$\end{tabular}  \\
$\gamma_u, \gamma_i \in \mathbb{R}^h$ & Learned $h$-dimension user/item embeddings.                                        \\
$\hat{x}_{u,i} \in \mathbb{R}$      &  The predicted score of item $i$ for user $u$.\\
$\hat{k}_{u,i} \in \mathbbm{1}^{|K|}$      &  Predicted justification (binary for all aspects). \\
$c^t_u \in \mathbb{R}^{|K|}$      &    \begin{tabular}[c]{@{}l@{}}Cumulative critique vector representing the\\user's evolving opinion about each aspect.\end{tabular}\\
$m^t_u \in \mathbbm{1}^{|K|}$      &   \begin{tabular}[c]{@{}l@{}}The user critique vector at turn $t$. $m^t_{u,a}$ is 1 if\\and only if the user critiqued $a$ at turn $t$.\end{tabular}\\ \bottomrule
\end{tabular}
\vspace{-0.4em}
\end{table}

\subsection{Conversational Recommendation}
\label{sec:litreview-critique}

Users often seek to make informed decisions around consumption, and controllability of a recommendation system is linked to improved user satisfaction \cite{controlsat}.
We thus turn to \textit{conversational recommendation} as a way to iteratively engage with a user to learn their preferences with the goal of recommending a suitable item \cite{convrec0}.
We view recommenders as domain experts engaging with human customers,
able to elicit user preferences and requirements and suggest appropriate items in the course of the conversation.

% ConvRec with implicit feedback
In early interactive recommender systems, users were only able to give binary ``like" and ``dislike" feedback without further explanation \cite{img0}.
One line of research used such feedback to refine the search space for retrieving desired images from the web \cite{img1}.
\citet{seeker} extended this approach to interactive product search.
More recently, multi-armed bandit approaches to conversational recommendation \cite{bandit1,bandit2} leverage exploration-exploitation algorithms to maximize the information learned from feedback at each turn.

% ConvRec as a Q&A game
Another line of work treats conversational recommendation as a form of task-oriented dialog where users express opinions about specific aspects of an item.
At each turn, the user is either a) asked if they prefer a specified aspect; or b) recommended an item \cite{convrec1}.
Self-supervised bot-play has been explored as a way to train such conversational dialog agents \cite{recdial,BotPlayRec}, but such approaches require Wizard-of-Oz style data \cite{WoZ} with humans playing the role of expert and seeker.
The quality of such dialog data depends heavily on the domain knowledge and competence of crowd-workers, which makes it unsuitable for complex domains.
\citet{saur} uses templated dialog forms and trains a model to ask about aspects that are most informative of the user's preferences.
However, this forces the user to answer yes/no questions and restricts their flexibility when giving feedback.
Instead, we explore \textit{conversational critiquing}, where a user is presented with items and justifications, and is able to give feedback regarding any aspect in the justification.

\subsection{Conversational Critiquing}

Critiquing systems aim to help users incrementally construct their preferences in a way that mimics how humans refine their preferences and constraints depending on conversation context \cite{contextpref}.
Early critiquing methods relied on constraint-based programming to iteratively shrink the search space of items as users provided more critiques \cite{constraint1}.

More recently, \citet{WuCEVNCF} introduced a critiquing model with justifications via a list of natural language aspects mined from user reviews.
In this setting, users are able to interact with any aspect present in the generated justification.
\citet{TREC} generate a single sentence of explanation alongside the set of aspects, but still require users to interact with the aspect set.
\citet{CEVAE} use a variational auto-encoder (VAE) \cite{VAE} in place of the collaborative filtering model, learning a bi-directional mapping function between user latent representations and aspects they have expressed in reviews.
Such models can generate high precision justifications but have shown poor multi-turn recommendation performance.

Latent Linear Critiquing (LLC) methods do not generate justifications and instead allow users to critique any aspect from the vocabulary \cite{LLC,LLCR}.
After training a matrix factorization model to predict ratings, these models then learn a linear regressor to recover user embeddings from their historical aspect usage frequency.
A linear programming problem is then solved to weight a user's critiques during each turn of the conversation, which we observe to take an order of magnitude longer than VAE-based methods and our own.
Furthermore, while LLC assumes that user preferences are fully explained by their review texts, recent studies have shown that this assumption may be unfounded \cite{reviewsinrec}.

In \Cref{tab:lit-comparison} we compare our approach in context of recent frameworks for training dialog agents for conversational recommendation and conversational critiquing agents.

\begin{figure}[t!]
\centerline{\includegraphics[width=0.9\linewidth]{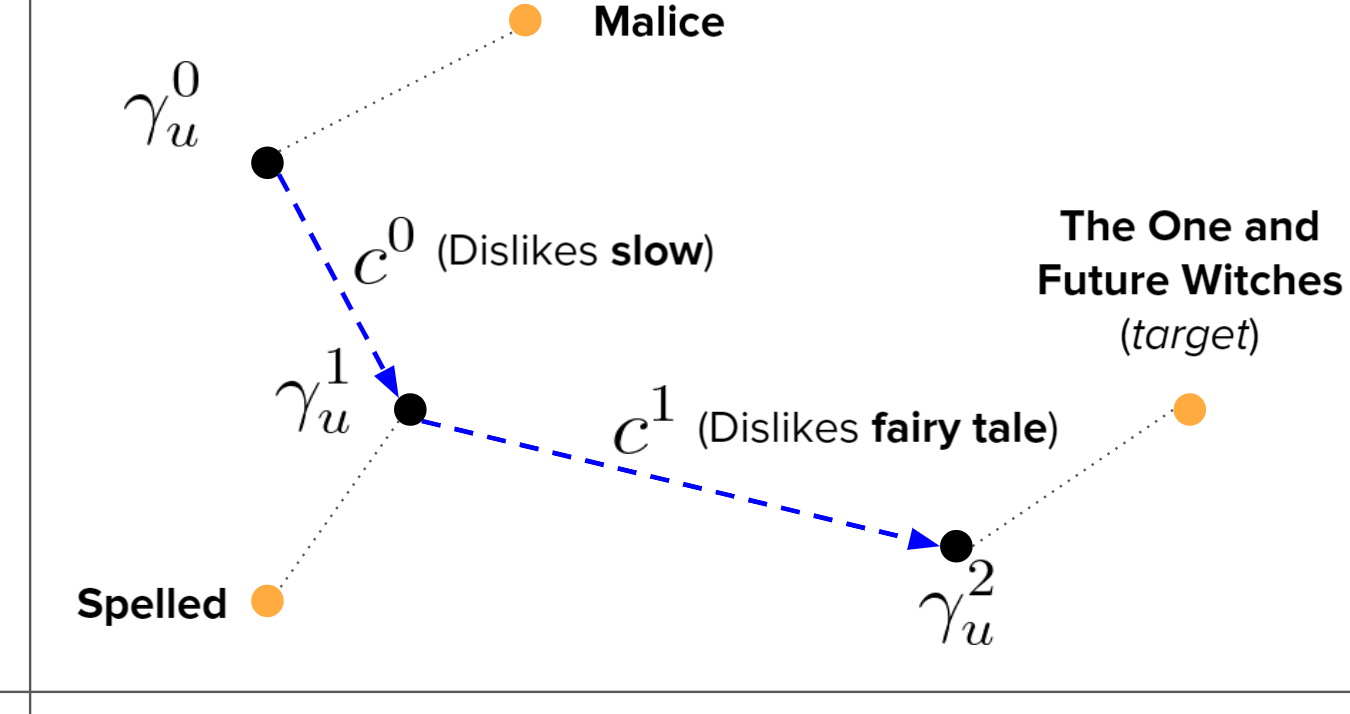}}
\caption{In (latent) conversational critiquing, user feedback about aspects ($c^0, c^1$) modifies our prior latent user preference vector $\gamma_u^0$ to bring it closer to the target item embedding.}
\label{fig:latent-critiquing}
\end{figure}

\section{Model}

\begin{figure}[t!]
\centerline{\includegraphics[width=0.95\linewidth]{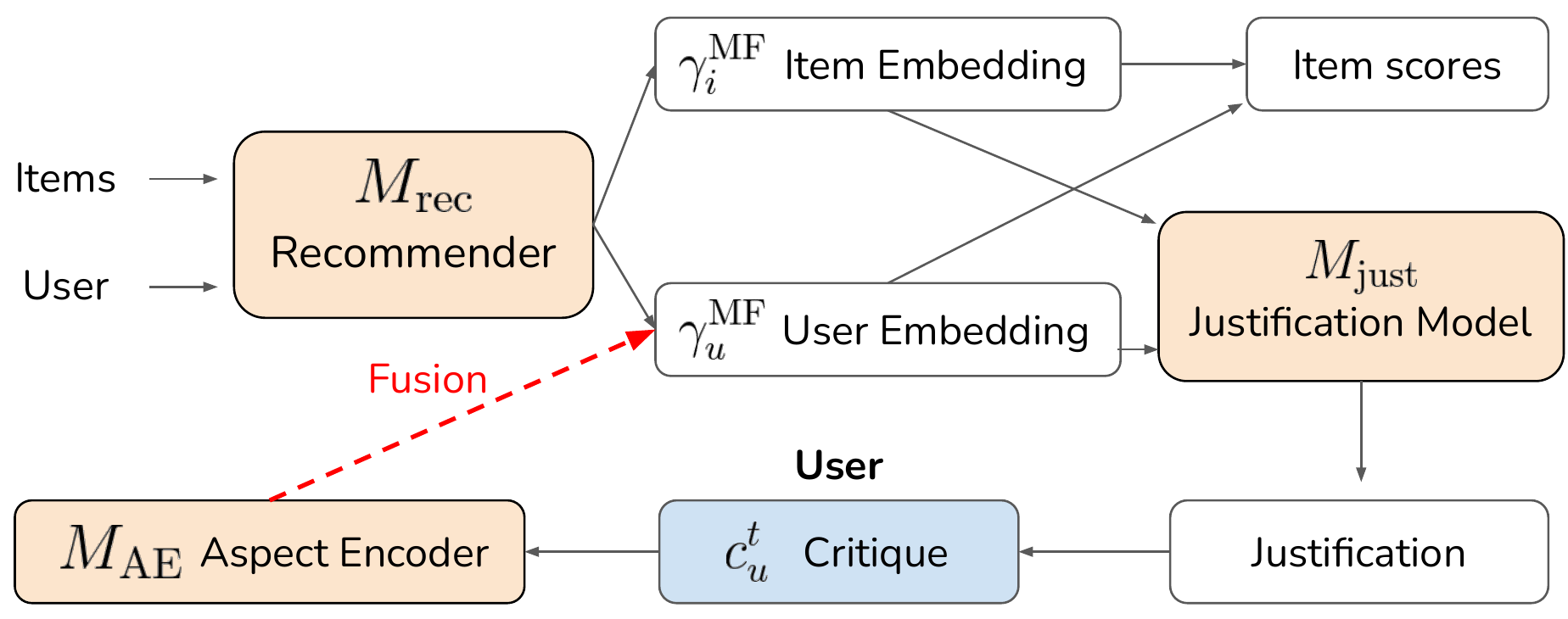}}
\caption{The proposed model architecture. Given a user, items, and critique vector, our model encodes the critique $M_{\text{AE}}(c^t_u)$ and fuses it with the user embedding $\gamma^{\text{MF}}_u$. The fused user representation $\gamma_u$ and item representation $\gamma_i$ are then used to predict the justification and score items.}
\label{fig:model}
\end{figure}

Our model consists of three sections, as seen in \Cref{fig:model}:
\begin{enumerate}
    \item A matrix factorization recommender model $M_{\text{rec}}$ that learns to embed users and items in an $h$-dimensional latent space;
    \item A justification head $M_{\text{just}}$ that predicts the aspects of an item toward which the user holds preferences;
    \item A critiquing function $f_\text{crit}$ that modifies a user's preference embedding based on user feedback about specific aspects.
\end{enumerate}

Our model supports multi-turn critiquing as shown in \Cref{fig:latent-critiquing}.
At each turn of a conversation, a user may provide explicit feedback about aspects they dislike about the current set of recommendations in the form of a critique ($c^t$).
The critiquing function $f_\text{crit}$ then uses this critique to modify our latent user representation $\gamma_u$ in order to bring it closer to the user's target item.

\subsection{Base Recommender System}
\label{sec:bpr}

Our method can be applied to any recommender system $M_{\text{rec}}$ that learns user and item representations.
We demonstrate its effectiveness using two popular methods based on matrix factorization and linear recommendation.

\emph{Bayesian Personalized Ranking} (BPR) \cite{bpr} is a matrix factorization recommender system that seeks to decompose the interaction matrix $\textbf{R} \in \mathbb{R}^{|U|\times|I|}$ into user and item representations \cite{matrixfactorization}.
BPR optimizes a ranked list of items given implicit feedback (a set of items with which a user has recorded a binary interaction).

We learn $h$-dimensional user and item embeddings ($\gamma_u^{\text{MF}}$, $\gamma_i^{\text{MF}}$), computing the score via the inner product: $\hat{x}_{u,i} = \langle \gamma_u^{\text{MF}}, \gamma_i^{\text{MF}} \rangle$.
At training time, the model is given a user $u$, observed item $i \in I_u^+$, and unobserved item $j \in I_u^-$.
We maximize the likelihood that the user prefers the observed item $i$ to the unobserved item $j$:
\begin{align}
    \mathcal{L}_R\ =\ P(i >_u j | \Theta)\ =\ \sigma(\hat{x}_{u,i} - \hat{x}_{u,j})
\end{align}
where $\sigma$ represents the sigmoid function $\frac{1}{1+e^{-x}}$.

\emph{Projected Linear Recommendation} (PLRec) is an SVD-based method to learn low-rank representations for users and items via linear regression \cite{plrec}.
The PLRec objective minimizes the following:
\begin{align}
    \label{eq:plrec}
    \argmin_W \sum_{u}\parallel r_{u} - r_{u}VW^T \parallel^2_2 + \Omega(W)
\end{align}
where $V$ is a fixed matrix obtained by taking a low-rank SVD approximation of $\mathbf{R}$ such that $\mathbf{R}=U \Sigma V^T$, and $W$ is a learned embedding matrix.
We thus obtain an $h$-dimensional user embedding $\gamma_u^{\text{MF}} = r_{u}V$ and $h$-dimensional item embedding $\gamma_i^{\text{MF}} = W_i$.

\subsection{Generating Justifications}

Our justification model $M_{\text{just}}$ consists of an aspect prediction head: a fully connected network with two $h$-dimensional hidden layers that predicts a score $s_{u,i,a}$ for each aspect $a$.
This model takes as input the sum of the learned user and item embeddings ($\gamma_u, \gamma_i$).

At training time, we incorporate an aspect prediction loss $\mathcal{L}_A$ by computing the binary cross entropy (BCE) for each aspect:
\begin{align}
    \mathcal{L}_A &= -\frac{1}{|A|} \sum_{a=0}^{|A|} \mathbf{k}^{I}_{i,a} \cdot \log p_{u,i,a} + (1 - \mathbf{k}^{I}_{i,a}) \cdot \log(1 - p_{u,i,a})
\end{align}
where $p_{u,i,a} = \sigma(s_{u,i,a})$ represents the likelihood of user $u$ caring about aspect $a$ in context of item $i$.
At inference time, we again compute the likelihood for each aspect ($p_{u,i,a} = \sigma(s_{u,i,a})$) and sample from the Bernoulli distribution with $p_{u,i,a}$ to determine which aspects $a$ appear in the justification.

\subsection{Encoding Aspects}
We posit that the user's latent representation can be partially explained by their written reviews.
Thus, we jointly learn an aspect encoder $M_{\text{AE}}$ alongside our recommendation model.
This takes the form of a linear projection from the aspect space to the user preference space:
$M_{\text{AE}}(c^t_u) = W^Tc^t_u + b$, where $c^t_u \in \mathbb{Z}^{|K|}$ is the critique vector representing the strength of a user's preference for each aspect.
We then fuse this aspect encoding with the latent user embedding from $M_{\text{rec}}$ to form the final user preference vector $\gamma_u$:
$\gamma_u = f(\gamma_u^{\text{MF}}, M_{\text{AE}}(c^t_u))$

For the BPR-based model, we use $f(a, b) = a + b$ as a fusion function, and for the PLRec-based model, we use $f(a, b) = \frac{a+b}{2}$.
During training, the aspect encoder takes in the user's aspect history: $c^t_u = \textbf{k}^U_u$.

\begin{algorithm}[t!]
\SetAlgoLined
Recommendation and Justification models $M_{\text{rec}}, M_{\text{just}}$\;
Critique fusion function $f_{\text{crit}}$\;
Seeker model $M_{\text{seeker}}$\;
\For{\upshape{each user} $u$}{
 \For{\upshape{goal item} $g \in I_u^+$ \upshape{(Evaluation set)}}{
    initialize loss $\mathcal{L}$\;
    initialize $\gamma^1_u$ from $M_{\text{rec}}$\;
    \For{\upshape{turn} $t \in range(1, T)$}{
        compute scores $\hat{x}^t_{u,i} = M_{\text{rec}}(\gamma^t_u, i)\ \forall\ i \in I$\;
        $\mathcal{L} \leftarrow \mathcal{L} + \delta^{t} \cdot \mathcal{L}_{\text{CE}}(g, \hat{x}^t_{u,i})$\;
        recommend item $\hat{i}^t = \argmax_i{\hat{x}^t_{u,i}}$\;
        \If{$\hat{i}^t = g$}{
            break session with success\;
        }
        generate justification $\hat{k}_{u,\hat{i}^t} = M_{\text{just}}(\gamma^t_u, \gamma_{\hat{i}^t})$\;
        $M_{\text{seeker}}$ critiques justification: $c^t_u$\;
        $\gamma^{t+1}_u \leftarrow f_{\text{crit}}(\gamma^t_u, c^t_u) $\;
    }
 }
}
\caption{Bot play framework for fine-tuning conversational recommenders.}
\label{alg:bot-play}
\end{algorithm}

\subsection{Training}
\label{sec:training}

To train our BPR-based model, we jointly optimize each component.
Each training example consists of a user $u$, an observed item $i \in I^+_u$ that the user has interacted with, and an unobserved item $j \in I^-_u$ that the user has not rated.
We predict scores for items $i$ and $j$:
\begin{align}
    \hat{x}_{u,i} = \langle \gamma_u, \gamma_i \rangle = \langle \gamma_u^\text{MF} + M_{\text{AE}}(\textbf{k}^U_u), \gamma_i \rangle
\end{align}
We first compute the BPR loss (see \Cref{sec:bpr}) with predicted scores $\hat{x}_{u,i}$ and $\hat{x}_{u,j}$.
We add the aspect prediction loss, scaled by a constant $\lambda_{\text{KP}}$ to the ranking loss for our training objective: $\mathcal{L} = \lambda_{\text{KP}} \mathcal{L}_A - \mathcal{L}_R$.
We find empirically that $\lambda_{\text{KP}}\in \{0.5, 1.0\}$ works well.

To train our PLRec-based model, we follow \citet{LLC} and separately optimize $M_\text{rec}$, $M_\text{just}$, and $M_\text{AE}$.
The user and item embeddings are learned via \cref{eq:plrec}.
We solve the following linear regression problem to optimize $M_{\text{AE}}$:
\begin{align}
    \argmin_{W, b} \sum_u \parallel \gamma^{\text{MF}}_u - M_\text{AE}(\mathbf{k}^U_u) \parallel^2_2 + \Omega(W)
\end{align}
Finally, we optimize the aspect prediction (justification) loss $\mathcal{L}_A$ to train the justification head.

\subsection{Conversational Critiquing with Our Models}
\label{sec:critiquing}

To perform conversational critiquing with a model trained using our framework, we adapt the latent critiquing formulation from \citet{LLC}, as shown in \Cref{fig:pipeline}.
Each conversation with a user $u$ consists of multiple turns.
At each turn $t$, the system assigns scores $\hat{x}^t_{u,i}$ for all candidate items $i$, and presents the user with the highest scoring item $\hat{i}$.
The system also justifies its prediction with a set of predicted aspects $\hat{k}^t_{u,i}$.
The user may either accept the recommended item (ending the conversation) or critique an aspect from the justification: $a\in \{a|\hat{k}_{u,i,a} = 1\}$.

Given a user critique, the system modifies the predicted scores for each item and presents the user with a new item and justification:
\begin{align}
    \hat{x}^{t+1}_{u,i} &= M_{\text{rec}}(\hat{\gamma}_u^{t+1}, i)\\
    \hat{k}^{t+1}_{u,i} &= M_{\text{just}}(\hat{\gamma}_u^{t+1}, i)\\
    \hat{\gamma}_u^{t+1} &\leftarrow f_{\text{crit}}(\hat{\gamma}_u^t, c^t_u)
\end{align}
Effectively, a user critique modifies our prior for the user's preferences; we then re-rank the items presented to the user.

At inference time, $c^t_u$ is the cumulative critique vector, initialized with the user's aspect history:
\begin{align}
    c^t_u &= c^{t-1}_u - \max(\textbf{k}^U_u, 1) \odot m^t_u; \quad c^0_u = \textbf{k}^U_u
\end{align}
where $\odot$ is element-wise multiplication.
We use $\max(\textbf{k}^U_u, 1)$ as the critique should match the strength of the user's previous opinion on the aspect---otherwise the encoding may have a small magnitude.
Even if a user has not mentioned an aspect in their previous reviews, the $\max$ ensures a non-zero effect from each critique.

\subsection{Learning to Critique via Bot Play}
\label{sec:bot-play}
We propose a framework for critiquing via bot play that simulates conversations when provided just a known set of user reviews.
We first pre-train our expert model (recommender model, justification model, and aspect encoder).
A seeker model $M_{\text{seeker}}$ is pre-trained via a simple user prior: when provided with a known target item and justification, it selects the most popular aspect present in the justification $\hat{k}_{u,\hat{i}}$ but not the target's historical aspects $\textbf{k}^I_{i}$ to critique.

For each training example (user $u$ and a goal item they have reviewed $g$), we allow the expert and seeker models to converse with the goal of recommending the goal item.
We fine-tune the expert by maximizing its reward (minimizing its loss) in the bot-play game (\Cref{alg:bot-play}).
We end the dialog after the goal item is recommended or a maximum session length of $T=10$ turns is reached.
We define the expert's loss as the cross entropy loss of recommendation scores per turn:
\begin{align}
    \mathcal{L}^{\text{expert}} &= \sum_t^{T} \delta^{t-1} \cdot \mathcal{L}_{\text{CE}}(g, \hat{x}^t_{u,i})
\end{align}
where $\delta$ is a discount factor\footnote{We use a discount factor of $\delta=0.9$} to encourage successfully recommending the goal item at earlier turns.
$\mathcal{L}_{\text{CE}}(g, \hat{x}^t_{u,i})$ is the cross entropy loss between predicted scores and the goal item:
\begin{align}
    \mathcal{L}_{\text{CE}}(g, \hat{x}^t_{u,i}) &= -\sum_{i\in I} P(i) \log_2 Q(i); \quad Q(i) = \frac{e^{\hat{x}^t_{u,i}}}{\sum^I_j e^{\hat{x}^t_{u,j}}}
\end{align}
where $P(i)$ is 1 if $g = i$ and 0 otherwise.
As the cross-entropy loss is continuous, we optimize the reward for each conversation ($u$, $i$).

\section{Experimental Setting}
To train our initial model, we select hyperparameters via AUC on the validation set.
We select hyperparameters for bot-play fine-tuning by evaluating the success rate at 1 (SR@1) on the validation set.
We train each model once, with three evaluation runs per experimental setting.
For baseline models, we re-used the authors' code.
We include additional training details in the supplementary materials.
\textbf{We will make our code available upon publication.}

\begin{table}[t!]
\centering
\caption{Descriptive statistics of datasets, including average unique aspects expressed in reviews per item and user.}
\label{tab:datasets}
\begin{tabular}{@{}lrrrrrr@{}}
\toprule
             & Users  & Items & Reviews & Asp. & Asp./Item & Asp./User \\ \midrule
Books        & 13,889 & 7,649 & 654,975 & 75  & 27.0  & 25.0 \\
Beer         & 6,369  & 4,000 & 935,524 & 75  & 60.2  & 54.6   \\
Music        & 5,635  & 4,352 & 119,081 & 80  & 20.0  & 16.5   \\ \bottomrule
\end{tabular}
\vspace{-0.5em}
\end{table}

\subsection{Datasets}
\label{sec:datasets}
We evaluate the quantitative performance of our model using three real-world, publicly available recommendation datasets: Goodreads Fantasy (Books) \cite{goodreads}, BeerAdvocate (Beer) \cite{beeradvocate}, and Amazon CDs \& Vinyl (Music) \cite{cdv1,cdv2}---each with over 100K reviews and ratings.
We keep only reviews with positive ratings, setting thresholds of $t > 4.0$ for Beer and Music and $t > 3.5$ for Books.
We partition each dataset into 50\% training, 20\% validation, and 30\% test splits.
Dataset statistics are shown in \Cref{tab:datasets}.

\subsection{Aspect Extraction}
\label{sec:aspect-extraction}
Our datasets do not contain pre-existing aspects, so we follow the pipeline of \cite{WuCEVNCF} to extract subjective aspects from user reviews:
\begin{enumerate}
    \item Extract lists of high-frequency unigrams and bigrams (nouns and adjective phrases only) from all user reviews;
    \item Prune the bigram keyphrase list using a Pointwise Mutual Information (PMI) threshold, ensuring aspects are statistically unlikely to have randomly co-occurred;
    \item Represent reviews as sparse binary vectors indicating whether each aspect was expressed in the review.
\end{enumerate}

Aspects describe a wide range of qualities; for beers, users commonly describe the malt (e.g.~roasted) and taste (e.g.~citrus).
For music, aspects range from perceived genres (e.g.~techno) to emotions (e.g.~soulful).
Users describe books by reacting to character descriptions (e.g.~strong female) and settings (e.g.~realistic)..

\begin{algorithm}[t!]
\SetAlgoLined
Recommendation and Justification models $M_{\text{rec}}, M_{\text{just}}$\;
Critique fusion function $f_{\text{crit}}$\;
\For{\upshape{each user} $u$}{
 \For{\upshape{goal item} $g \in I_u^+$ \upshape{(Evaluation set)}}{
    initialize $\gamma^1_u$ from $M_{\text{rec}}$\;
    \For{\upshape{turn} $t \in range(1, T)$}{
        compute scores $\hat{x}^t_{u,i} = M_{\text{rec}}(\gamma^t_u, i)\ \forall\ i \in I$\;
        recommend item $\hat{i}^t = \argmax_i{\hat{x}^t_{u,i}}$\;
        generate justification $\hat{k}^t_{u,\hat{i}} = M_{\text{just}}(\gamma^t_u, \gamma_{\hat{i}^t})$\;
        \If{$\hat{i}^t = g$}{
            break session with success\;
        }
        user critiques justification: $c^t_u$\;
        $\gamma^{t+1}_u \leftarrow f_{\text{crit}}(\gamma^t_u, c^t_u) $\;
    }
 }
}
\Return{\upshape{average success rate \& length}}
\caption{User Simulation Evaluation}
\label{alg:user-sim}
\end{algorithm}

\subsection{Multi-Step Critiquing}
\label{sec:user-sim}

Following prior work on conversational critiquing \cite{LLC,LLCR}, we simulate multi-step recommendation dialogs to assess model performance.
We randomly sample 500 user-item interactions from the test set to conduct user simulations following \Cref{alg:user-sim} for each user $u$ and goal item $g$.
At each turn, we recommend an item $\hat{i}^t$ to the user alongside a set of aspects $\hat{k}_{u, \hat{i}^t}$.
If the goal item is not recommended, the user will critique an aspect $a$ from the justification that is inconsistent with the goal item aspects: $a \in \{a\ |\ \hat{k}^t_{u,\hat{i}^t,a}=1\ \&\ \textbf{k}^I_{g,a}=0\}$
We set a maximum session limit of $T=10$ turns.

To evaluate how our models behave with different user behaviors, we simulate each observation with three different critique selection strategies \cite{LLCR}:
\begin{itemize}
    \item \textbf{Random}: We assume the user randomly chooses an aspect. This assumes no prior knowledge on the part of the user.
    \item \textbf{Pop}: We assume the user selects the most popular aspect used across all training reviews.
    \item \textbf{Diff}: We assume the user selects the aspect that deviates most from the goal item reviews. In simulations, we select the aspect with the largest frequency differential between the goal item and current item: $\argmax_a (\textbf{k}^I_{\hat{i}^t,a} - \textbf{k}^I_{g,a})$ 
\end{itemize}

In all critiquing settings, a user may not critique the same aspect multiple times in a session, and any recommended items are removed from consideration in the following turns.

\begin{figure}[t!]
\centerline{\includegraphics[width=0.95\linewidth]{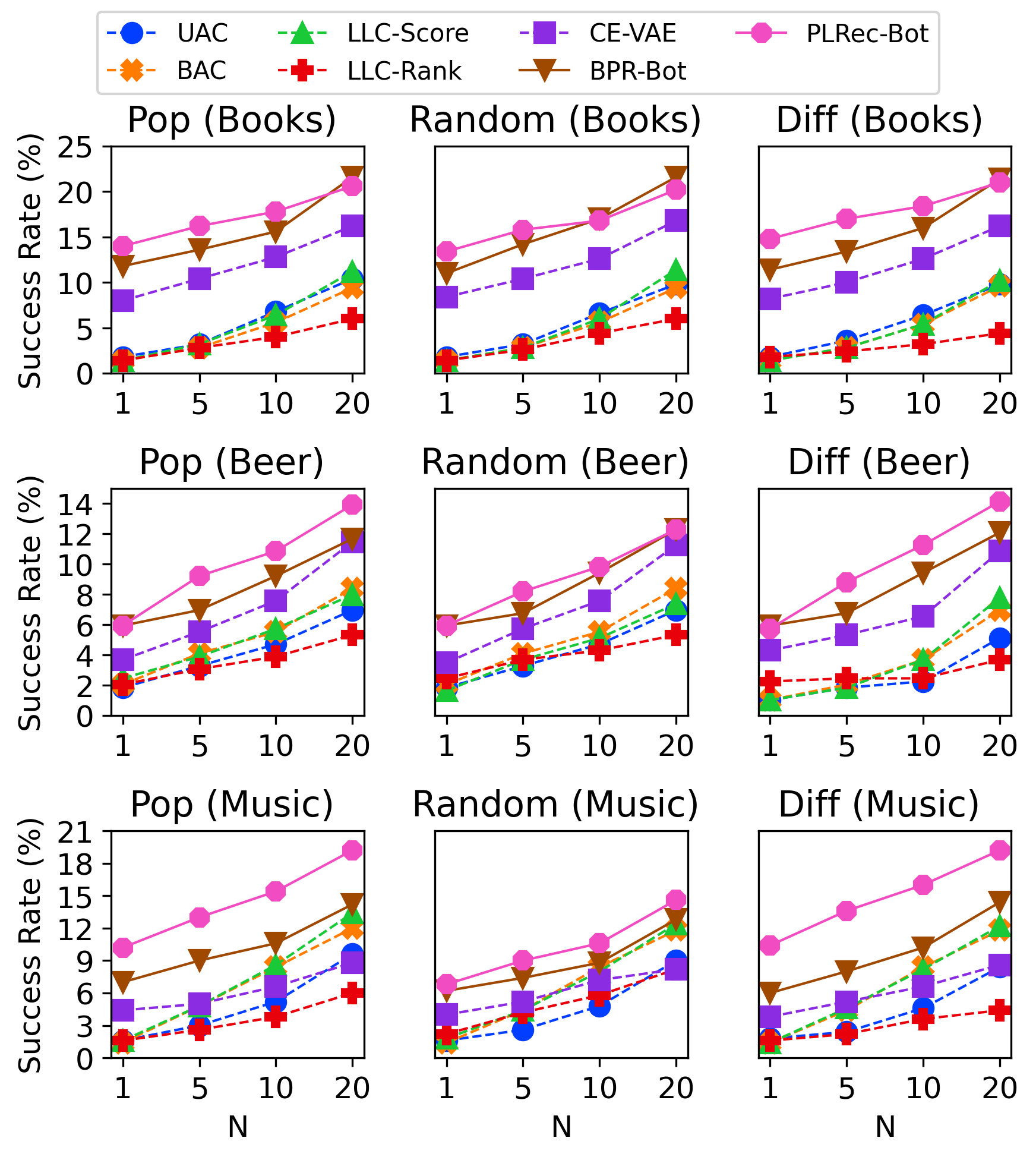}}
\vspace{-0.8em}
\caption{Success Rate @ N (\% dialogs where target item rank $\leq$ N) across datasets and user models. BPR-Bot (brown triangle) and PLRec-Bot (pink circle) out-perform baselines (dashed) in all settings.}
\label{fig:results-base}
\vspace{-1.0em}
\end{figure}

\subsection{Candidate Algorithms}
\label{sec:base_models}

As our method can be applied to any base recommender system $M_{\text{rec}}$, we apply our framework to train models based on BPR and PLRec (see \Cref{sec:bpr})---\textbf{BPR-Bot} and \textbf{PLRec-Bot}, respectively.

We assess Latent Linear Critiquing (LLC) baselines, which embed critique vectors $c^t_u$ in the same $h$-dimensional space as the latent user representation $\gamma_u$.
$f_\text{crit}$ is defined as a weighted sum of the embedding for each critiqued aspect, alongside the original user preference vector.
\textbf{UAC} \cite{LLC} averages the initial user embedding and all critiqued aspect embeddings.
\textbf{BAC} \cite{LLC} first averages critiqued aspects, and then averages the result with the initial user embedding.
\textbf{LLC-Score} \cite{LLC} learns the weights via a linear program maximizing the posterior rating differences between items containing critiqued aspects and those without.
Instead of directly optimizing the scoring margin, \textbf{LLC-Rank} \cite{LLCR} minimizes the number of ranking violations.
These models cannot generate justifications; we binarize the historical aspect frequency vector for the item ($\textbf{k}^I_{u,\hat{i}^t}$) as a justification at each turn.
We compare against these models to evaluate whether generating personalized justifications can improve critiquing.

We also compare against a state-of-the-art variational conversational recommender, \textbf{CE-VAE} \cite{CEVAE}---an improvement on the \citet{WuCEVNCF} justified critiquing model---which jointly learns to recommend and justify.
CE-VAE learns a VAE with a bidirectional mapping between critique vectors and the user latent preference space.
We compare our models to CE-VAE to assess how justification quality impacts multi-turn critiquing performance.

\section{Experiments}

\begin{figure}[t!]
\centerline{\includegraphics[width=0.95\linewidth]{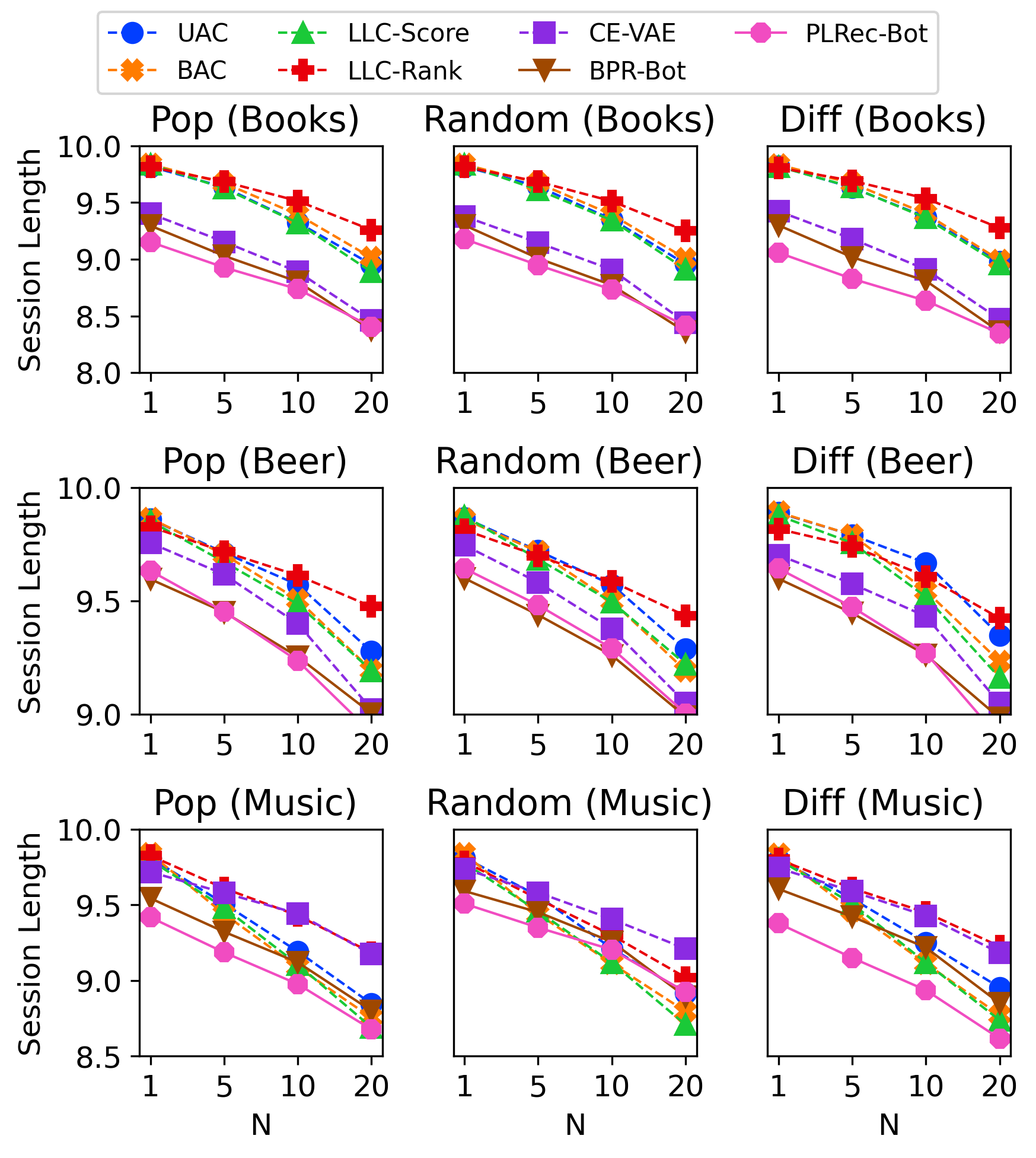}}
\vspace{-0.8em}
\caption{Avg. number of turns before the target item reaches rank N, across datasets and user models. BPR-Bot (brown triangle) and PLRec-Bot (pink circle) promote target items faster than baselines (dashed), especially for low N.}
\label{fig:results-base-turns}
\vspace{-0.3em}
\end{figure}

\newpara{RQ1: Can our framework enable multi-step critiquing?}
To measure multi-step critiquing performance, we assess the average success rate and session length following \citet{LLC}.
Success rate measures the percentage of sessions in which the target item reaches rank N, and session length measures the average length of sessions with a limit of 10 iterations.
Success rates and session lengths for each dataset and user behavior model are shown in \Cref{fig:results-base} and \Cref{fig:results-base-turns}, respectively.

\vfill

% Prior in bot play
Our models are fine-tuned via bot-play with a seeker model that assumes one particular user behavior: popularity-based critique selection.
As such, we expect it to perform better in the Pop user setting.
However, BPR-Bot and PLRec-Bot succeeds at a higher rate in fewer turns than baselines under \textit{all} user settings---including random aspect critiquing, which assumes no prior on user behavior.

\vfill

% VAE Baselines
\emph{Variational Baseline.}
Despite its strong first-turn recommendation performance and high-fidelity justifications, CE-VAE is out-performed by our models in all nine settings across all metrics.
This supports our observation that the training method to learn a bi-directional mapping between latent user preferences and a justification causes a trade-off between justification quality and critiquing ability.

\vfill

% Linear baselines
\emph{Linear Baselines.}
We further observe that linear critiquing models (UAC, BAC, LLC-Score, and LLC-Rank) perform poorly on multi-step critiquing, especially when trying to find the goal item outright ($N=1$).
This confirms our observation that the method of co-embedding aspect critiques with learned user latent preferences ignores the existence of user preferences not explained by review text.
This additionally suggests that generating personalized justifications helps users more effectively choose aspects to critique.

\vfill

\begin{figure}[t!]
\centerline{\includegraphics[width=1.0\linewidth]{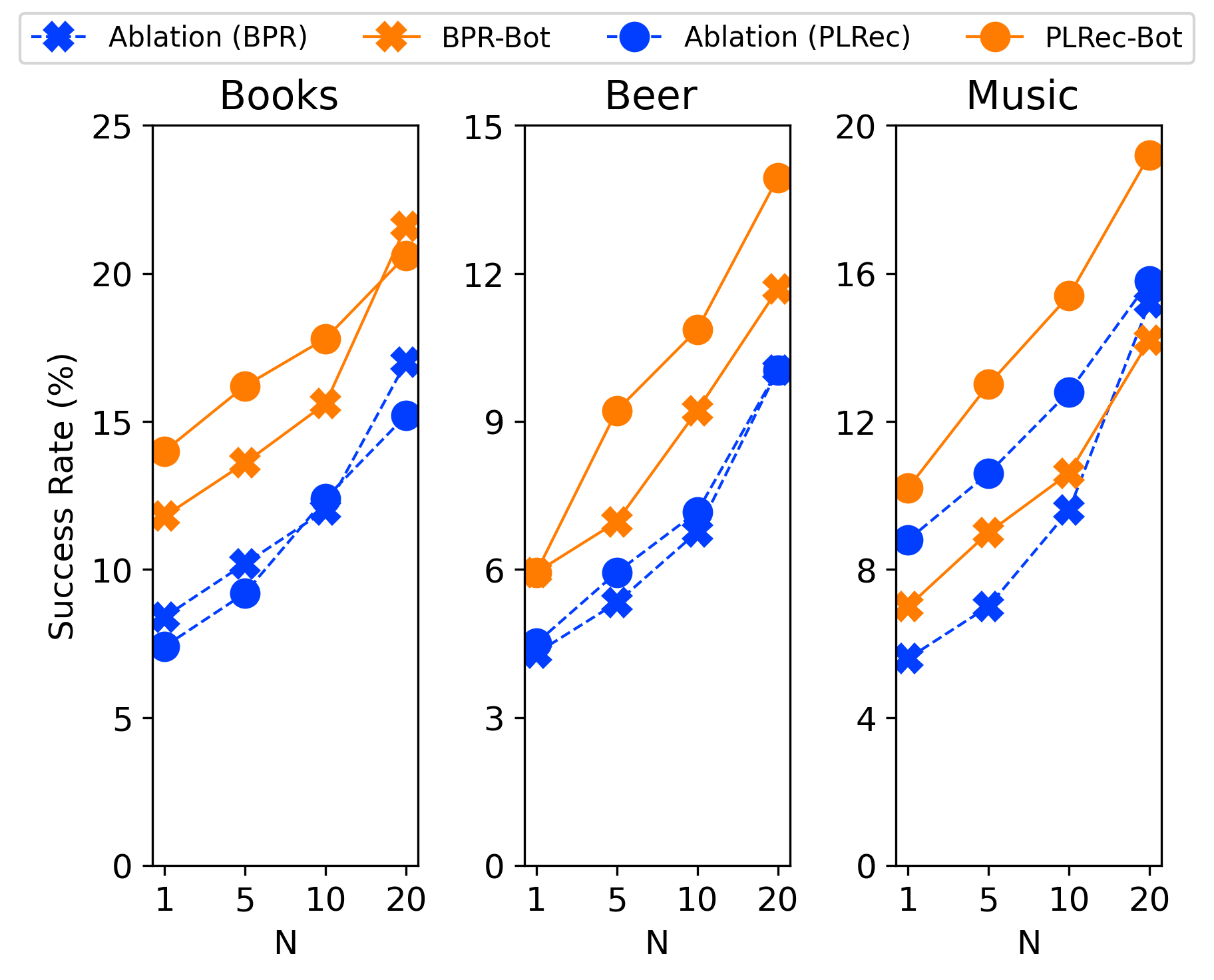}}
\vspace{-0.5em}
\caption{Success Rate @ N (\% dialogs where target item rank $\leq$ N), comparing bot-play methods (orange) against non-bot-play ablations (blue). Bot-play fine-tuning improves target item ranking across datasets compared to the ablation, for both BPR-Bot (crosses) and PLRec-Bot (circles).}
\label{fig:ablation}
\vspace{-0.3em}
\end{figure}

% Overall observations
In general, the large item space makes it difficult for critiquing models to reach the goal item within the turn limit, with the best model reaching the goal item in only 6-15\% of sessions.
This suggests that practical conversational critiquing systems may benefit from constraint-based filtering as well as starting the session from an initial set of user requirements---while users rarely enter a conversation knowing their full preference set \cite{prefdiscovery}, they often start with a limited set of broad requirements (e.g.~when buying a car, they want an SUV or a coupe).
We demonstrate in RQ3 that our model can be combined with constraint-based query refinement to quickly reach significantly higher success rates.

\vfill

\newpara{RQ2: Can our bot-play framework improve multi-step critiquing performance?}
We next compare BPR-Bot (left) and PLRec-Bot (right) in \Cref{fig:ablation} against ablated versions that were trained using the first step of our framework but \emph{not} fine-tuned via bot-play.
For clarity, we display only results using the Pop user behavioral model, as we observe the same trends with the Random and Diff user models.
In domains with relatively high aspect occurrence across reviews (Books, Beer), we observe that bot-play confers a 3-6\% improvement in success rate for various N.
This demonstrates that we can effectively train conversational recommender systems using our bot-play framework using domains with rich user reviews in lieu of crowd-sourced dialog transcripts.

\vfill

\begin{figure}[t!]
\centerline{\includegraphics[width=1.0\linewidth]{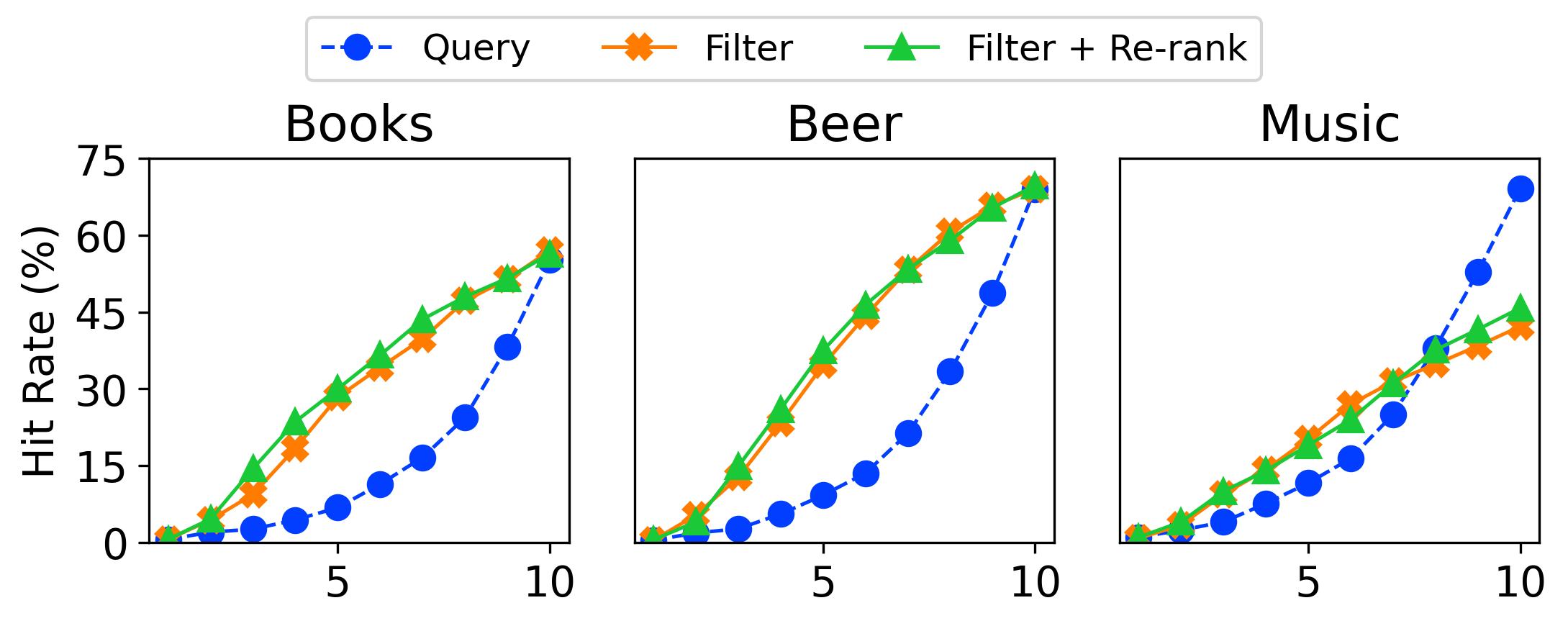}}
\vspace{-0.8em}
\caption{Hit rate by turn for query refinement models on each dataset with multi-step critiquing up to 10 turns.}
\label{fig:results-HR-filter}
\end{figure}

In domains with more sparse coverage of subjective aspects (i.e.~Music), we observe lower improvement when using bot-play.
Here, our model may not encounter sufficient examples of rare aspects being critiqued.
In future work, we will explore methods to add noise to our user model to ensure that the bot-play process encounters more rare aspects.
We will also investigate additional losses for bot-play, including ranking losses instead of cross entropy.

\vfill

We confirm that our method is model-agnostic, as it improves conversational recommendation success rates for both the matrix factorization-based (BPR) and linear (PLRec) recommender systems.
We also observe that models with a higher latent dimensionality ($h \in [50, 400]$ for PLRec-Bot vs. $h = 10$ for BPR-Bot) benefit more from bot-play, suggesting that our method effectively learns to navigate complex user preference spaces.

% \medskip
\newpara{RQ3: Can our models be effectively combined with query refinement?}
\label{sec:filtering}
So far, we have assumed that users provide \emph{soft} feedback via critiques: even if a user has critiqued aspect $a$ during a session, future suggested items may still contain aspect $a$.
This assumption holds for some aspects: for example, even if previous users mentioned that a song was dispassionate, a user may find it emotional and enjoyable.
However, in a real-world setting that user may reject the suggestion after reading the reviews.
Thus, we experiment with treating critiques as hard feedback: if a user critiques some aspect $a$, we prune all candidate items whose reviews mention $a$.
We compare three models in this setting, with the turn-0 ranked list of candidate items initialized from BPR-Bot.

\vfill

The \textbf{Query} baseline model suggests one item per turn and asks the user whether they like aspect $a$.
If the user answers yes, we prune all candidate items whose reviews have not expressed $a$: $I^{t+1} \leftarrow \{i\ \forall\ i \in I^{t} | \textbf{k}^I_{i,a} = 1\}$.
Otherwise, we prune all candidates whose reviews have expressed the aspect: $I^{t+1} \leftarrow \{i\ \forall\ i \in I^{t} | \textbf{k}^I_{i,a} = 0\}$.
At each turn, we pick the aspect that most evenly divides the remaining candidate items: $\argmin_a \vert |I^+_a| - |I^-_a| \vert$
Effectively, we perform binary search over our candidate space, and expect to find the target item within $\log{|I|}$ turns.

\vfill

In the \textbf{Filter} model, we suggest an item alongside a generated justification per turn.
When a user critiques aspect $a$, we prune candidate items whose reviews have expressed $a$: $I^{t+1} \leftarrow \{i\ \forall\ i \in I^{t} | \textbf{k}^I_{i,a} = 0\}$.
We extend this model via our learned critiquing function $f_{\text{crit}}$ to further modify the user preference vector and re-compute scores for the remaining items.
This hybrid \textbf{Filter+Re-rank} model then re-ranks the remaining candidate items for the next turn.
We conduct user simulations with the Pop user model following \Cref{alg:user-sim}, and plot the success rate by turn---rate of achieving the goal item $g$ at or before turn $t$---in \Cref{fig:results-HR-filter}.

Binary queries are guaranteed to eventually find the answer, but the queried aspect may not be related to suggested items.
By allowing the user to provide negative critiques, we can rapidly reduce the search space at early turns.
Across domains the success rate rises much faster in the first 6-10 turns for Filter and Filter+Re-rank compared to binary querying.
Re-ranking after filtering improves performance across domains, suggesting that we have learned how user critiques relate to their latent preferences for other aspects.

\begin{table}[t!]
\centering
\caption{Conversation-level human evaluation via ACUTE-EVAL. Win (W) and Loss (L) percentages are reported while ties are not. All results statistically significant with $p < 0.05$.}
\label{tab:human-eval-bpr}
\resizebox{\linewidth}{!}{%
\begin{tabular}{lcccccccc}
\toprule
\bf BPR-Bot  & \multicolumn{2}{c}{Useful} & \multicolumn{2}{c}{Informative} & \multicolumn{2}{c}{Knowledgeable} & \multicolumn{2}{c}{Adaptive} \\ 
vs       & W            & L            & W              & L             & W               & L              & W             & L            \\ \midrule
Ablation & \bf 78           & 10           & \bf 73             & 11            & \bf 68              & 15             & \bf 85            & 5            \\
CE-VAE   & \bf 83           & 9            & \bf 74             & 10            & \bf 63              & 16             & \bf 81            & 8            \\ \midrule\midrule
\bf PLRec-Bot  & \multicolumn{2}{c}{Useful} & \multicolumn{2}{c}{Informative} & \multicolumn{2}{c}{Knowledgeable} & \multicolumn{2}{c}{Adaptive} \\ 
vs       & W            & L            & W              & L             & W               & L              & W             & L            \\ \midrule
Ablation & \bf 86           & 5           & \bf 78             & 7            & \bf 74              & 8             & \bf 81            & 9            \\
CE-VAE   & \bf 87           & 7            & \bf 79             & 11            & \bf 77              & 12             & \bf 83            & 10            \\ \bottomrule
\end{tabular}%
}
\end{table}

% Domain effects
For the Beer and Books domains, the filtering approach reaches higher success rates compared to binary querying same high success rate within the session turn limit (70.7\% vs. 69.7\% and 57.0\% vs. 55.2\%, respectively).
We see less of a benefit in the Music domain.
Relative aspect sparsity may play a role:
per \Cref{tab:datasets}, only 25\% of possible aspects are expressed for the average item.
There also exists a longer tail of aspects expressed only for a small set of items in Music compared to the other datasets.
As such, user critiques prune fewer candidate items on average in Music.

% Extensible framework
Our bot-play framework can be easily adapted to train models incorporating hard critiquing constraints by pruning candidate items.
One possible extension involves masking the cross entropy (fine-tuning) loss to only adjust the scores of non-pruned items, setting pruned item scores to a large negative value: $\hat{x}_{u,i} = -1e15\ \forall\ i \in I^+_a$.
We also wish to explore fine-tuning with a ranking loss during bot-play, to encourage the model to rank items containing a critiqued aspect $i \in I^+_a$ below those without.

\section{Human Study}
\label{sec:user-study}

\newpara{Human Evaluation}
To assess the quality of the simulated conversations during bot-play, we conduct human evaluations with 100 samples.
Following ACUTE-EVAL \cite{acuteeval}, we conduct a comparative evaluation of each sample conversation on four criteria: which agent seems more useful, informative, knowledgeable and adaptive.
We compare each bot-play model (\textbf{BPR-Bot} and \textbf{PLRec-Bot}) against an ablative version (with no bot-play fine-tuning) and the best baseline model (CE-VAE).

Each sample is evaluated by three annotators.
We observe substantial \cite{fleiss_interp} inter-annotator agreement, with Fleiss Kappa \cite{fleisskappa} of 0.67, 0.79, 0.73, and 0.60 for the usefulness, informativeness, knowledgeable, and adaptiveness criteria, respectively.

\begin{table}[t!]
\centering
\caption{Turn- and conversation-level feedback from cold-start user study. Statistically significant results are \textbf{bolded}.}
\label{tab:ustudy-results}
\begin{tabular}{@{}l|rrr|r@{}}
\toprule& Useful     & Informative  & Adaptive  & Would use    \\ \midrule
Ablation & 0.67$\pm$0.24  & 0.75$\pm$0.21    & 0.64$\pm$0.27 & 41\%  \\
BPR-Bot & \textbf{0.79$\pm$0.24} & \textbf{0.88$\pm$0.18} & \textbf{0.78$\pm$0.23} & \textbf{69\%} \\ \bottomrule
\end{tabular}
\end{table}

BPR-Bot and PLRec-Bot are judged to be significantly more informative and knowledgeable compared to ablative models and CE-VAE, showing that our justification module accurately predicts aspects of a recommended item.
We design the usefulness and adaptiveness criteria to capture how our framework aids the user in achieving their conversational goal (i.e.~recommending the most relevant item within a minimum number of turns).
Compared to the alternatives, models trained under our bot-play framework are judged to be more useful and adapt their recommendations in a manner more consistent with critiques.

Our framework allows us to train conversational agents that are useful and engaging for human users: evaluators overwhelmingly judged the models trained via bot-play to be more useful, informative, knowledgeable, and adaptive compared to CE-VAE and ablated variants.

\newpara{Cold-Start User Study}
We conduct a user study using items and reviews from the Books dataset to evaluate our model's ability to provide useful conversational recommendations in real-time.
We recruited 32 real human users to interact with our \textbf{BPR-Bot} recommender and another 32 to interact with the ablation model (no conversational fine-tuning).
As evaluators do not correspond to users in our training data, we initialize each conversation with the average of all learned latent user representations.

\vfill

At each turn, the user is presented with the three top-ranked items and their justifications (list of aspects), and is allowed to critique multiple aspects.
On average, users critiqued two aspects per turn---this suggests that when training conversational models we should assume multiple critiques at each turn.

\vfill

We evaluate our systems following \citet{acuteeval}: at each turn, we ask our users if the generated justifications are \emph{informative}, \emph{useful} in helping to make a decision, and whether our system \emph{adapted} its suggestions in response to the user's feedback.
We provide four options for each question: yes, weak-yes, weak-no, and no, mapping these values to a score between 0 and 1 \cite{DBLP:journals/corr/abs-2105-03761}.
We display the normalized aggregated score for each question in \Cref{tab:ustudy-results}.
We find that \textbf{BPR-Bot} significantly out-scores the ablation model in all three metrics ($p < 0.01$), showing that fine-tuning our model on a bot-play framework instills a stronger ability to respond to techniques and provide meaningful justifications---even for unseen users.

At the end of a conversation, we additionally ask the user how frequently (if at all) they would choose to engage with our conversational agent in their daily life.
69\% of users indicated they would ``often" or ``always" use BPR-Bot to find books, compared to 41\% of users for the ablation model.
We thus find that fine-tuning our model via bot-play also makes it significantly ($p < 0.05$) more useful for new users.

\section{Conclusion}
In this work, we aim to develop conversational agents for recommendation that engage with users following common modes of human dialog: justifying why suggestions were made and incorporating feedback about certain aspects of an item to provide better recommendations at the next turn.
We present a framework for training conversational recommenders in this modality via self-supervised bot-play.
Our framework is model-agnostic and allows conversational recommenders to be trained on any domain with review data.
We use two popular underlying recommender systems to train the \textbf{BPR-Bot} and \textbf{PLRec-Bot} conversational agents using our framework, demonstrating quantitatively on three datasets that our models 1) offer superior multi-turn recommendation performance compared to current state-of-the-art methods; 2) can be effectively combined with query refinement techniques to quickly converge on suitable items; and 3) can iteratively refine suggestions in real-time, as shown in user studies.
In future work, we aim to adapt our framework to natural language critiques (i.e.~complete utterances), allowing users to freely express their feedback in a less restrictive way.

\bibliographystyle{ACM-Reference-Format}
\bibliography{www2022}
\newpage

\appendix

\section{Additional Training Details}
All experiments were conducted on a machine with a 2.2GHz 40-core CPU, 132GB memory and one RTX 2080Ti GPU.
We use PyTorch version 1.4.0
and optimize our models using the Rectified Adam \cite{radam} optimizer.
Best hyperparameters for each base recommender system model are shown in \Cref{tab:hparams}.

\section{Time Complexity}

% Time taken
In \Cref{tab:timings}, we report the mean and standard error of time taken per turn for LLC-Score, CE-VAE, BPR-Bot, and PLRec-Bot.
As baseline code does not leverage the GPU, we also critique with PLRec-Bot and BPR-Bot on the CPU only.
We observe LLC-Score and PLRec-Bot to be an order of magnitude slower per critiquing cycle compared to CE-VAE and BPR-Bot.

BPR-Bot shows acceptable latency for real-world applications (sub-10 ms), and we observe empirically in our cold-start user study (\Cref{sec:user-study}) that we can host BPR-Bot as a real-time recommendation service.
Time trials were conducted with batch size of 1; production throughput 
% and in production settings throughput 
can be improved further with parallel processing.
% via larger batches.
Each model executes using a different framework (\texttt{numpy} for LLC-Score, \texttt{Tensorflow} for CE-VAE, and \texttt{Pytorch} for PLRec-Bot/BPR-Bot), which may contribute to differences in inference speed.

\section{User Evaluation}

\begin{figure}[t!]
\centerline{\includegraphics[width=0.95\linewidth]{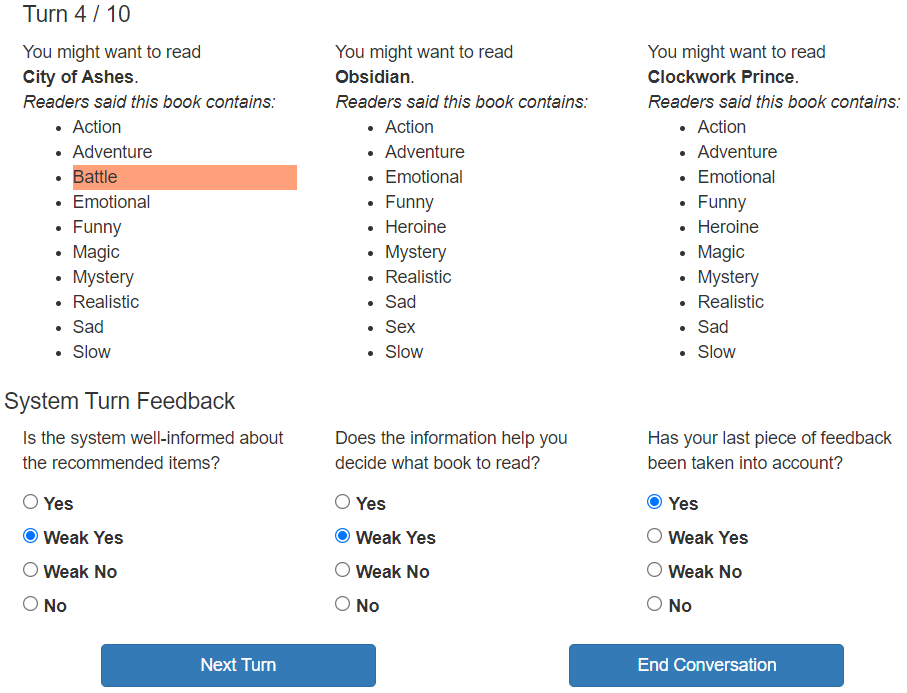}}
\caption{User interface for user study, with turn-level feedback prompts and an example of a critiqued aspect (``Battle")}
\label{fig:ustudy-ui}
\end{figure}

An image of the interface used for our cold-start user study (\Cref{sec:user-study}) is shown in \Cref{fig:ustudy-ui}.

\begin{table}[t!]
\centering
\caption{Mean and standard error of wall-clock time (ms) per turn of critiquing for linear (LLC-Score) and variational (CE-VAE) baselines vs. our models (BPR-Bot, BPR-PLRec)}
\label{tab:timings}
\begin{tabular}{@{}lrrrr@{}}
\toprule
      & LLC-Score            & CE-VAE             & BPR-Bot   & PLRec-Bot\\ \midrule
Books & 40.64 $\pm$ 20.46 & 4.61 $\pm$ 1.16 & 2.70 $\pm$ 3.95 & 48.84 $\pm$ 14.08\\
Beer  & 15.94 $\pm$ 14.52 & 3.26 $\pm$ 1.18 & 2.54 $\pm$ 2.36 & 49.43 $\pm$ 14.81\\
Music & 42.21 $\pm$ 21.04 & 3.36 $\pm$ 1.37 & 2.25 $\pm$ 0.62 & 6.80 $\pm$ 7.53\\ \bottomrule
\end{tabular}
\end{table}

\begin{table*}[h!]
\centering
\caption{Best hyperparameter settings for each base recommendation model. UAC, BAC, LLC-Score, LLC-Rank models use PLRec as a base model. BPR-Bot uses BPR as a base model.}
\label{tab:hparams}
\begin{tabular}{@{}llllllllll@{}}
\toprule
Dataset & Model  & $h$   & LR     & $\lambda_{\text{L2}}$ & $\lambda_{\text{KP}}$ & $\lambda_c$ & $\beta$   & Epoch & Dropout \\ \midrule
Books   & BPR    & 10  & 0.001  & 0.01       & 0.5        & --        & --     & 200   & --      \\
        & PLRec  & 50  & --     & 80         & --         & --        & --     & 10    & --      \\
        & CE-VAE & 100 & 0.0001 & 0.0001     & 0.01       & 0.01      & 0.001  & 300   & 0.5     \\ \midrule
Beer    & BPR    & 10  & 0.001  & 0.01       & 0.5        & --        & --     & 200   & --      \\
        & PLRec  & 50  & --     & 80         & --         & --        & --     & 10    & --      \\
        & CE-VAE & 100 & 0.0001 & 0.0001     & 0.01       & 0.01      & 0.001  & 300   & 0.5     \\ \midrule
Music   & BPR    & 10  & 0.01   & 0.1        & 1.0        & --        & --     & 200   & --      \\
        & PLRec  & 400 & --     & 1000       & --         & --        & --     & 10    & --      \\
        & CE-VAE & 200 & 0.0001 & 0.0001     & 0.001      & 0.001     & 0.0001 & 600   & 0.5     \\ \bottomrule
\end{tabular}
\end{table*}

\end{document}